\documentclass[twoside,11pt]{article}

\usepackage{graphicx}
\usepackage{epstopdf}
\usepackage{upgreek}
\usepackage{url}

\newcommand{\m}{\mathbf{m}}

\begin{document}

\title{Photonic Delay Systems as Machine Learning Implementations}

\author{Michiel Hermans\thanks{OPERA Photonique, Universit/'e Libre de Bruxelles, Avenue F. Roosevelt 50, 1050 Brussels (michiel.hermans@ulb.ac.be)},
\and  Miguel C. Soriano\thanks{Instituto de F\'isica Interdisciplinar y Sistemas Complejos, IFISC (UIB-CSIC), Campus Universitat de les Illes Balears, E-07122 Palma de Mallorca, Spain}, 
\and Joni Dambre\thanks{ELIS departement, Ghent University, Sint Pietersnieuwstraat 41, 9000 Ghent, Belgium}, 
\and Peter Bienstman\thanks{INTEC departement, Ghent University, Sint Pietersnieuwstraat 41, 9000 Ghent, Belgium}, 
\and Ingo Fischer\thanks{Instituto de F\'isica Interdisciplinar y Sistemas Complejos, IFISC (UIB-CSIC), Campus Universitat de les Illes Balears, E-07122 Palma de Mallorca, Spain}}

\maketitle

\begin{abstract}
Nonlinear photonic delay systems present interesting implementation platforms for machine learning models. They can be extremely fast, offer great degrees of parallelism and potentially consume far less power than digital processors. So far they have been successfully employed for signal processing using the Reservoir Computing paradigm. In this paper we show that their range of applicability can be greatly extended if we use gradient descent with backpropagation through time on a model of the system to optimize the input encoding of such systems. We perform physical experiments that demonstrate that the obtained input encodings work well in reality, and we show that optimized systems perform significantly better than the common Reservoir Computing approach. The results presented here demonstrate that common gradient descent techniques from machine learning may well be applicable on physical neuro-inspired analog computers.
\end{abstract}

\section{Introduction}
Applied research in neural networks is currently strongly influenced by available computer architectures. Most strikingly, the increasing availability of general-purpose graphical processing unit (GPGPU) programming has sped up the computations required for training (deep) neural networks by an order of magnitude. This development allowed researchers to dramatically scale up their models, in turn leading to the major improvements on state-of-the-art performances on tasks such as computer vision (\cite{Krizhevsky2012,Ciresan2010}).\\
One class of neural models which has only seen limited effects of the boost in speed from GPUs are recurrent models. Recurrent neural networks (RNNs) are very interesting for processing time series, as they can take into account an arbitrarily long context of their input history. This has important implications in tasks such as natural language processing, where the desired output of the system may depend on context that has been presented to the network a relatively long time ago. In common feedforward networks such dependencies are very hard to include without scaling up the model to an impractically large size. Recurrent networks, however, can--at least in principle--carry along relevant context as they are being updated.\\
In practice, recurrent models suffer from two important drawbacks. First of all, where feedforward networks fully benefit from massively parallel architectures in terms of scalability, recurrent networks, with their inherently sequential nature do not fit so well into this framework. Even though GPUs have been used to speed up training RNNs (\cite{Sutskever2011,Hermans2013}), the total obtainable acceleration for a given GPU architecture will still be limited by the number of sequential operations required in an RNN, which is typically much higher than in common neural networks. The second issue is that training RNNs is a notoriously slow process due to problems associated with fading gradients, which is especially cumbersome if the network needs to learn long-term dependencies within the input time series. Recent attempts to solve this problem using the Hessian-free approach have proved promising (\cite{Martens2011}). Other attempts using stochastic gradient descent combined with more heuristic ideas have been described in \cite{Bengio2013}.\\
In this paper we will consider a radical alternative to common, digitally implemented RNNs. A steadily growing branch of research is concerned with \emph{Reservoir Computing} (RC), a concept which employs high-dimensional, randomly initialized dynamical systems (termed the \emph{reservoir}) to perform feature extraction on time series (\cite{Jaeger2001,Jaeger2004,Maass2002a,Steil2004,Lukosevicius2009}). Despite its simplicity, RC has several important advantages over traditional  gradient descent training methods. First of all, the training process is extremely fast. Only output weights are trained, and this is performed by solving a single linear system of equations. Second, and of great importance, the RC concept is applicable to \emph{any} non-linear dynamical system, as long as it exhibits consistent responses, a high-dimensional state space, and fading memory. This has opened lines of research that go beyond common digital implementations and into analog physical implementations. The RC concept has been demonstrated to work on a variety of physical implementation platforms, such as water ripples (\cite{Fernando2003}), mechanical constructs and tensegrity structures (\cite{Caluwaerts2013,Hauser2011}), electro-optical devices (\cite{Larger2012,Paquot2012}), fully optical devices (\cite{Brunner2013}) and nanophotonic circuits (\cite{Vandoorne2008, Vandoorne2014}). As opposed to digital implementations, physical systems can offer great speed-ups, inherent massive parallelism, and great reductions in power consumption. In this sense, physical dynamical systems as machine learning implementation platforms may one day break important barriers in terms of scalability. In the near future, especially optical computing devices might find applications in several tasks where fast processing is essential, such as in optical header recognition, optical signal recovery, or fast control loops.\\
The RC paradigm, despite its notable successes, still suffers from an important drawback. Its inherently unoptimized nature makes it relatively inefficient for many important machine learning problems. When the dimensionality of the input time series is low, the expansion into a high-dimensional nonlinear space offered by the reservoir will provide a sufficiently diverse set of features to approximate the desired output. If the input dimensionality becomes larger, however, relying on random features becomes increasingly difficult as the space of possible features becomes so massive. Here, optimization with gradient descent still has an important edge over the RC concept: it can shape the necessary nonlinear features automatically from the data.\\
In this paper we aim to integrate the concept of gradient descent in neural networks with physically implemented analog machine learning models. Specifically, we will employ a physical dynamical system that has been studied extensively from the RC paradigm, a delayed feedback electro-optical system (\cite{Larger2012,Paquot2012,Soriano2013}). In order to use such a system as a reservoir, an input time series is encoded into a continuous time signal and subsequently used to drive the dynamics of the physical setup. The response of the device is recorded and converted to a high-dimensional feature set, which in turn is used with linear regression in the common RC setup. In this particular case, the randomness of RC is incorporated in the input encoding. This encoding is performed offline on a computer, but is usually completely random. Even though efforts have been performed to improve this encoding in a generic way (by ensuring a high diversity in the network's response (\cite{Rodan2011, Appeltant2014})), a way to create task-specific input encodings is still lacking.\\
In \cite{Hermans2014}, the possibility to use \emph{backpropagation through time} (BPTT) (\cite{Rumelhart1986}) as a generic optimization tool for physical dynamical systems was addressed. It was found that BPTT can be used to find remarkably intricate solutions to complicated problems in dynamical system design. In \cite{Hermans2014a} simulated results of BPTT used as an optimization method for input encoding in the physical system described above were presented. In this paper we go beyond this work and show for the first time experimental evidence that model-based BPTT is a viable training strategy for physical dynamical systems. We choose two often-used high-dimensional datasets for validation, and we show that input encoding that is optimized using BPTT in a common machine learning approach, provides a significant boost in performance for these tasks when compared to random input encodings. This not only demonstrates that machine learning approaches are more broadly applicable than is generally assumed, but also that physical analog computers can in fact be considered as parametrizable machine learning models, and may play a significant role in the next generation of signal processing hardware.\\
This paper is structured as follows: first of all we discuss the physical system and its corresponding model in detail. We explain how we convert the continuous-time dynamics of the system into a discrete-time update equation which we use as model in our simulation. Next, we present and analyze the results on the tasks we considered and compare experimental and simulated results. 
\section{Physical System}\label{section:Physical_System}
\begin{figure}
\centering
\includegraphics[width=0.6\textwidth]{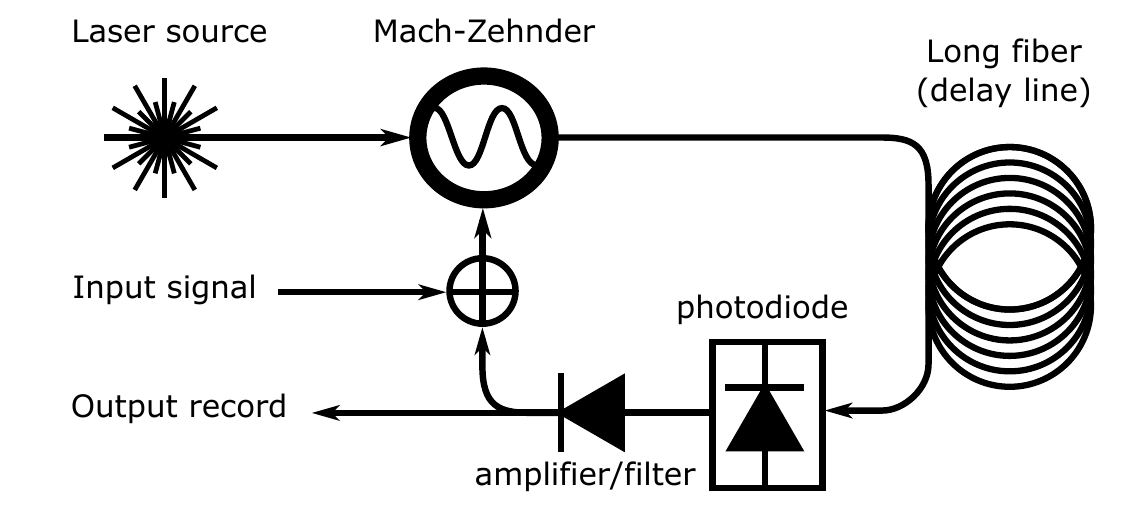}
\caption{Schematic depiction of a delay-coupled Mach-Zehnder interferometer.}\label{fig:DCMZ}
\end{figure}

In this section we will explain the details of the physical system. We will start by formally introducing its delay dynamics operating in continuous time. Next, we will explain how the feedback delay can be used for realizing a high-dimensional state space encoded in time, and we demonstrate that--combined with special input and output encoding--the setup can be seen as a special case of RNN. Finally we explain how we discretize the system's input and output encoding, which enables us to approximate the dynamics of the system by a discrete-time update  equation.\\
The physical system we employ in this paper is a delayed feedback system exhibiting Ikeda-type dynamics (\cite{Larger2004,Weicker2012}). We provide a schematic depiction of the physical setup in Figure \ref{fig:DCMZ}. It consists of a laser source, a Mach-Zehnder modulator, a long optical fiber ($\approx 4 \textrm{ km}$) which acts as a physical delay line, and an electronic circuit which transforms the optical beam intensity in the fiber into a voltage. This voltage is amplified and low-pass filtered and can be measured to serve as the system output. Moreover, it is added to an external input voltage signal, and then serves as the driving signal for the Mach-Zehnder modulator. The measured output signal is well described by the following differential equation (\cite{Larger2012}):
\begin{equation}
T\dot{a}(t) = -a(t) + \beta\left[\sin^2(a(t-D) + z(t) + \phi) - 1/2\right].\label{eq:diff_eq}
\end{equation}
Here, the signal $a(t)$ corresponds to a measured voltage signal (down to a constant scaling and bias factor). The factor $T$ is the time scale of the low-pass filtering operation in the electronic circuit, equal to 0.241 $\upmu$s, $\beta$ is the total amplification in the loop, which in the experiments can be varied by changing the power of the laser source. $D$ is the delay of the system, which has been chosen as 20.82 $\upmu$s. $z(t)$ is the external input signal, and $\phi$ is a constant offset phase (which can be controlled by setting a bias voltage), which we set at $\pi/4$ for all results presented in this paper. For ease of notation we will call the system a \emph{delay-coupled Mach-Zehnder}, which we abbreviate as DCMZ.\\
Note that the parameters $\beta$ and $\phi$, together with the global scaling of the input signal $z(t)$, control the global dynamical behavior of the system (\cite{Larger2012}). Indeed, previous research in the RC context have identified the role of these parameters in connection with task performance. They found that good performance is usually found when the parameters put the system in an asymptotically stable regime. For instance, if we keep $\phi=\pi/4$, and $\beta<1$, the system state will always fall back to zero in the absence of input. In the case of $\beta>1$, the state of the system will spontaneously start to oscillate, which has a detrimental effect on task performance. In this paper we will simply use values for $\beta$ and $\phi$ that were found to generally work well in the reservoir setup. 
\subsection{Input and Output Encoding}\label{section:IO}
\begin{figure}
\centering
\includegraphics[width=0.9\textwidth]{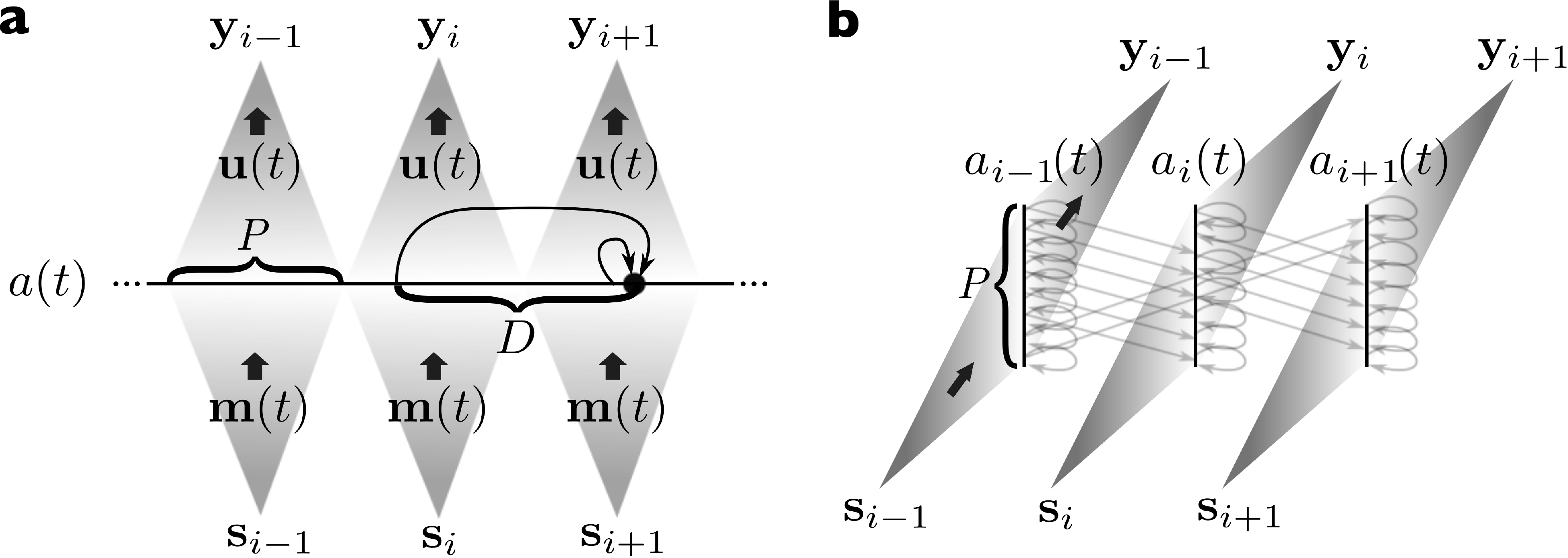}
\caption{Schematic representation of the masking principle. \textbf{a}: Depiction of the input time series $\mathbf{s}_i$ and the way it is converted into a continuous-time signal by means of the input masking signals $\m(t)$. The horizontal line in the middle shows the time evolution of the system state $a(t)$. We have depicted two connection arrows at one point in time, which indicate that $a(t)$ depends on its immediately preceding value (due to the low-pass filtering operation), and its delayed value. The state trajectories are divided into segments each of which are projected to an output instance $\mathbf{y}_i$. \textbf{b}: The same picture as in panel \textbf{a}, but now represented as a time-unfolded RNN. We have shown the connections between the states as light grey arrows, but note that there are in principle infinitely many connections.) }\label{fig:maskRNN}
\end{figure}

Delay-coupled systems have--in principle--an infinite-dimensional state space, as these systems directly depend on their full history covering an interval of one delay time. This property has been the initial motivation for using delay-coupled systems in the RC paradigm in the past years. Suppose we have a multivariate input time series, which we will denote by $\mathbf{s}_i$, for $i \in \{1,2,\cdots,S\}$, $S$ being the total number of instances (the length of the input sequence). Each $\mathbf{s}_i$ is a column vector of size $N_\mathrm{in}\times1$, with $N_\mathrm{in}$ the number of input dimensions. We wish to construct an accompanying output time series $\mathbf{y}_i$. We convert each data point $\mathbf{s}_i$ to a continuous-time segment $z_i(t)$ as follows:
\[z_i(t) = m_0(t) + \m^{\textsf{T}}(t)\mathbf{s}_i,\]
where $m_0(t)$ and $\m(t)$ are \emph{masking signals}, which are defined for $t\in\left[0\cdots P\right]$, with $P$ the masking period. The signal $m_0(t)$ is scalar, and constitutes a bias signal, and $\m(t)$ is a column vector of size $N_\textrm{in}\times1$. The total input signal $z(t)$ is then constructed by time-concatenation of the segments $z_i(t)$:
\[z(t) = z_i(t \textrm{ mod } P)\;\;\;\;\textrm{for}\;\;\;\;t\in\{(i-1)P\cdots iP\}.\]
Similarly, we define an output mask $\mathbf{u}(t)$. We divide the state variable time traces $a(t)$ in segments $a_i(t)$ of duration $P$ such that 
\[a(t) = a_i(t \textrm{ mod } P)\;\;\;\;\textrm{for}\;\;\;\;t\in\{(i-1)P\cdots iP\}.\]
The output time series $\mathbf{y}_i$ is then defined as
\begin{equation}
\mathbf{y}_i= \mathbf{y}_0 + \int_0^P{dt\;a_i(t)\mathbf{u}(t)}.\label{eq:outp}
\end{equation}
It is possible to see the delay-coupled dynamical system combined with the masking principle as a special case of an infinite-dimensional discrete-time recurrent neural network, as illustrated in Figure \ref{fig:maskRNN}. The recurrent weights, connecting the hidden states over time, are fixed, and manifested by the delayed feedback connection. The input and output weights correspond to the input and output masks. 
\\
The role of the parameters $D$ and $P$ is important to consider. If they are equal to each other the recurrent network analogy, as shown in Figure \ref{fig:maskRNN}b, reduces to a network where all nodes have self-connections, and interaction between different nodes between different tasking periods is due to a combination of the low-pass filtering effect and the self-connection. If the difference between $D$ and $P$ is small, there will be direct time-interaction between different nodes. In fact, using a difference of one masking step between $D$ and $P$ has been the basis for opto-electronic systems that do not have a low-pass filter (\cite{Paquot2012}). If the difference between $D$ and $P$ becomes significant it is difficult to anticipate how performance will be affected. If $D\ll P$, most interactions will happen within a single masking period, such that there will be little useful interaction between the nodes at different time steps. If $D\gg P$, the nodes interact over connections that bridge several time steps. We found that, for small differences of $D$ and $P$, there is little to no noticeable effect on performance, such that we kept $D=P$, as was used in previous publications.
\\
In practice, we cannot measure the state trajectory with infinite time resolution, nor can we produce signals with an arbitrary time dependency, as there will always be constraints that limit the maximum bandwidth of the generated signals. Therefore, we assume that $m_0(t)$, $\m(t)$ and $\mathbf{u}(t)$ all consist of piecewise constant signals\footnote{Note that with a finite frequency bandwidth we cannot produce immediate jumps from one constant level to the next. Therefore, we make sure that the duration of each constant part is much longer than the transient in between, and we can safely ignore it.}, which are segmented in $N_m$ parts, $N_m$ being the number of \emph{masking steps}:
\[m_0(t) = m_{0k}\;\;\;\;\textrm{for}\;\;\;\;t\in\{(k-1)P_m\cdots kP_m\},\]
\[\m(t) = \m_{k}\;\;\;\;\textrm{for}\;\;\;\;t\in\{(k-1)P_m\cdots kP_m\},\]
\begin{equation}
\mathbf{u}(t) = \mathbf{u}_k\;\;\;\;\textrm{for}\;\;\;\;t\in\{(k-1)P_m\cdots kP_m\},\label{eq:PW}
\end{equation}
where the length of each step is given by $P_m = P/N_m$. This means that we now have a finite number of parameters that fully determine $m_0(t)$, $\m(t)$ and $\mathbf{u}(t)$. Note that, due to our choice of $P=D$, $P_m$ will by definition be an integer number of times the delay length $D$. This is convenient for the next section, where we will make a discrete-time approximation of the system, but it is not a necessary requirement of the system to perform well. 
\subsection{Converting the System to a Trainable Machine Learning Model}
In \cite{Hermans2014} it was shown that BPTT can be applied to models of continuous-time dynamical systems. Indeed, it is perfectly possible to simulate the system using differential equation solvers and consequently compute parameter gradients. One issue, however, is the significant computational cost. Note that, in a common discrete-time RNN, a single state update corresponds to a single matrix-vector multiplication and the application of a nonlinearity. In our case it involves the sequential computation of the full time trace of $a_i(t)$. This is considerably more costly to compute, especially given the fact that--as in most gradient descent algorithms--we may need to compute it on large amounts of data and this for multiple thousands of iterations.\\
Due to the piecewise constant definition of $\mathbf{u}(t)$ we can make a good approximation of $a(t)$. First we combine Equations \ref{eq:outp} and \ref{eq:PW}. This gives us:
\[\mathbf{y}_i = \sum_{k=1}^{N_m} \int_{(k-1)P_m}^{kP_m}{dt\;\mathbf{u}_ka_i(t)} = \sum_{k=1}^{N_m} \mathbf{u}_k\bar{a}_{ik},\]
where $\bar{a}_{ik} =  \int_{(k-1)P_m}^{kP_m}{dt\;a_i(t)}$. This means that we can represent $a_i(t)$ by a finite set of variables $\bar{a}_{ik}$. To represent the full time trace of $a(t)$ we adopt a simplified notation as follows\footnote{Please do not confuse with the index $i$ in $a_i(t)$. Here the index indicates single masking steps, rather than full mask periods.}: $\bar{a}_j = \bar{a}_{ik}$, where $j = (i-1)N_m + k$.\\
Now we make the following approximation: we assume that for the duration of a single masking step, we can replace the term $a(t-D)$ by $\bar{a}_{i-N_m}$, that is, we consider it to be constant. With this assumption, we can solve Equation \ref{eq:diff_eq} for the duration of one masking step:
\begin{equation}
a(t) = \gamma_i + \left(\hat{a}_i - \gamma_i\right)\exp\left(-\frac{t}{T}\right)\;\;\;\;\;\;\textrm{ for }\;\;\;\;t\in\{0\cdots P_m\}, \label{eq:a_ms} 
\end{equation}
with 
\[\gamma_i = \beta\left[\sin^2(\bar{a}_{i-N_m} + z(t) + \phi) - 1/2\right],\]
and $\hat{a}_i$ the value of $a(t)$ at the start of the interval. Integrating over the interval $t = \{0\cdots P_m\}$ we find:
\[\bar{a}_i = (\hat{a}_i - \gamma_i)\kappa + P_m\gamma_i,\]
with $\kappa = 1-\textrm{e}^{-P_m/T}$. We can eliminate $\hat{a}_i$ as follows. First we derive from Equation \ref{eq:a_ms} that $\hat{a}_{i+1} = (\hat{a}_i - \gamma_i)\textrm{e}^{-P_m/T} + \gamma_i$. If we combine this expression with the following two:
\[\bar{a}_i = (\hat{a}_i - \gamma_i)\kappa + P_m\gamma_i,\]
\[\bar{a}_{i+1} = (\hat{a}_{i+1} - \gamma_{i+1})\kappa + P_m\gamma_{i+1},\]
we can eliminate $\hat{a}_i$, and we end up with the following update equation for $\bar{a}_{i}$:
\[
\bar{a}_{i+1} = \rho_o \bar{a}_i + \rho_1\gamma_i + \rho_2\gamma_{i+1},
\]
with $\rho_0 = \textrm{e}^{-P_m/T}$, $\rho_1 = T\kappa - P_m\textrm{e}^{-P_m/T}$, and $\rho_2 = P_m-T\kappa$. This leads to a relatively quick-to-compute update equation to simulate the system. BPTT can also be readily applied on this formula, as it is a simple update equation just like for a common RNN. This is the simulation model we used for training the input and output masks of the system.\\
We verified the accuracy of this approximation both on measured data of the DCMZ and on a highly accurate simulation of the system. For the parameters used in the DCMZ we got very good correspondence with the model (obtaining a correlation coefficient between simulated and measured signals of 99.6\%).
\subsection{Hybrid Training Approach}
One challenge we faced when trying to match the model with the experimentally measured data was that we obtained a sufficiently good correspondence only when we very carefully fitted the values for $\beta$ and $\phi$. We can physically control these parameters, but exactly setting their numerical values turned out not to be trivial in the experiments, especially since they tend to show slight drifting behavior over longer periods of time (in the order of hours). As a consequence, it turned out to be a challenge to train parameters in simulation, and simply apply them directly on the DCMZ. Therefore, we applied a hybrid approach between gradient descent and the RC approach. We train both the input and output masks in simulations. Next, we only use the input masks for the physical setup. After recording all the data, we retrained the output weights using gradient descent, this time on the measured data itself. The idea is that the input encoding will produce highly useful features for the system even when it is trained on a model that may show small, systematic differences with the physical setup. 
\subsection{Input Limitations}
One additional physical constraint is the fact that the voltages that can be generated by the electronic part of the system are limited within a range set by its supply voltage. The output voltage of the electronic part serves as the input of the Mach-Zehnder interferometer, and corresponds to the term $a(t-D) + z(t)$ in the argument of the squared sine in Equation \ref{eq:diff_eq} (the offset phase $\phi$ is controlled by a separate voltage source). The voltage range we were able to cover before the amplifiers started to saturate, roughly corresponded to  a range of $\left[-\pi/2\cdots \pi/2\right]$ in Equation \ref{eq:diff_eq}: one full wavelength. Instead of accounting for the saturation of the amplifiers in our simulations, we made sure that when the input argument $z(t)$ went outside of this range, we mapped it back into this range by adding or subtracting $\pi$. Note that this has no effect on Equation \ref{eq:diff_eq} due to the periodicity of the squared sine. Due to the addition of the input signal with the delayed feedback $a(t-D)$, there is still a chance that the total argument falls out of the range $\left[-\pi/2\cdots \pi/2\right]$, but in practice such occurrences turned out to be rare, and could safely be ignored.  
\section{Experiments}
\begin{figure}
\centering
\includegraphics[width=1\textwidth]{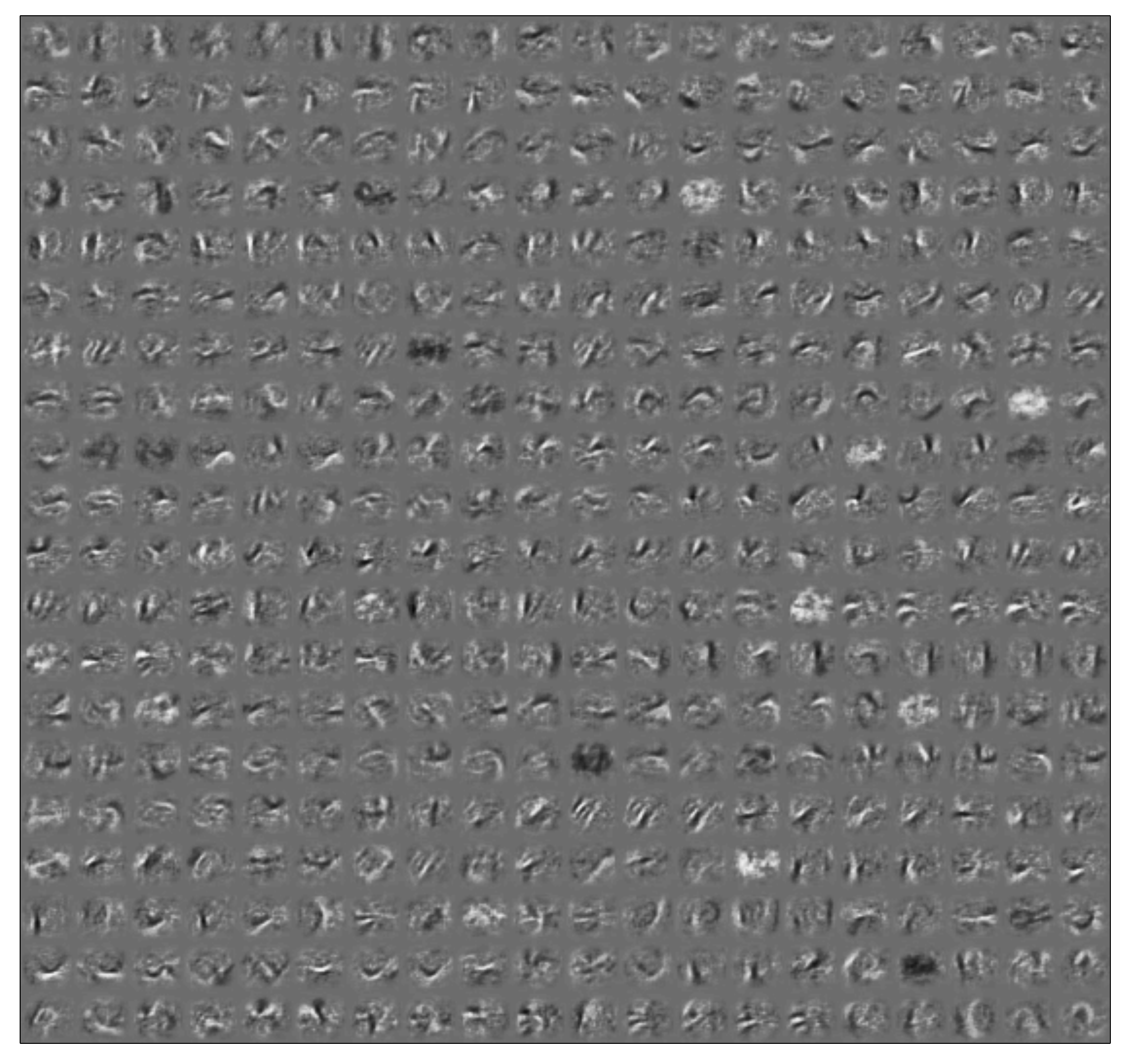}
\caption{Depiction of the image features present in the input masks for the MNIST task. We have shown the input weights of the 400 masking steps, which we have reshaped into a 20$\times$20 grid of 28$\times$28 pixel representations, corresponding to the receptive fields of each masking step (which can be considered virtual "neurons"). Time (progression of the masking steps, and hence physical time) runs row by row. Notice that the order in which they occur is not random, but rather similar features are grouped in time.}\label{fig:MNIST_features}
\end{figure}

\begin{figure}
\centering
\includegraphics[width=1\textwidth]{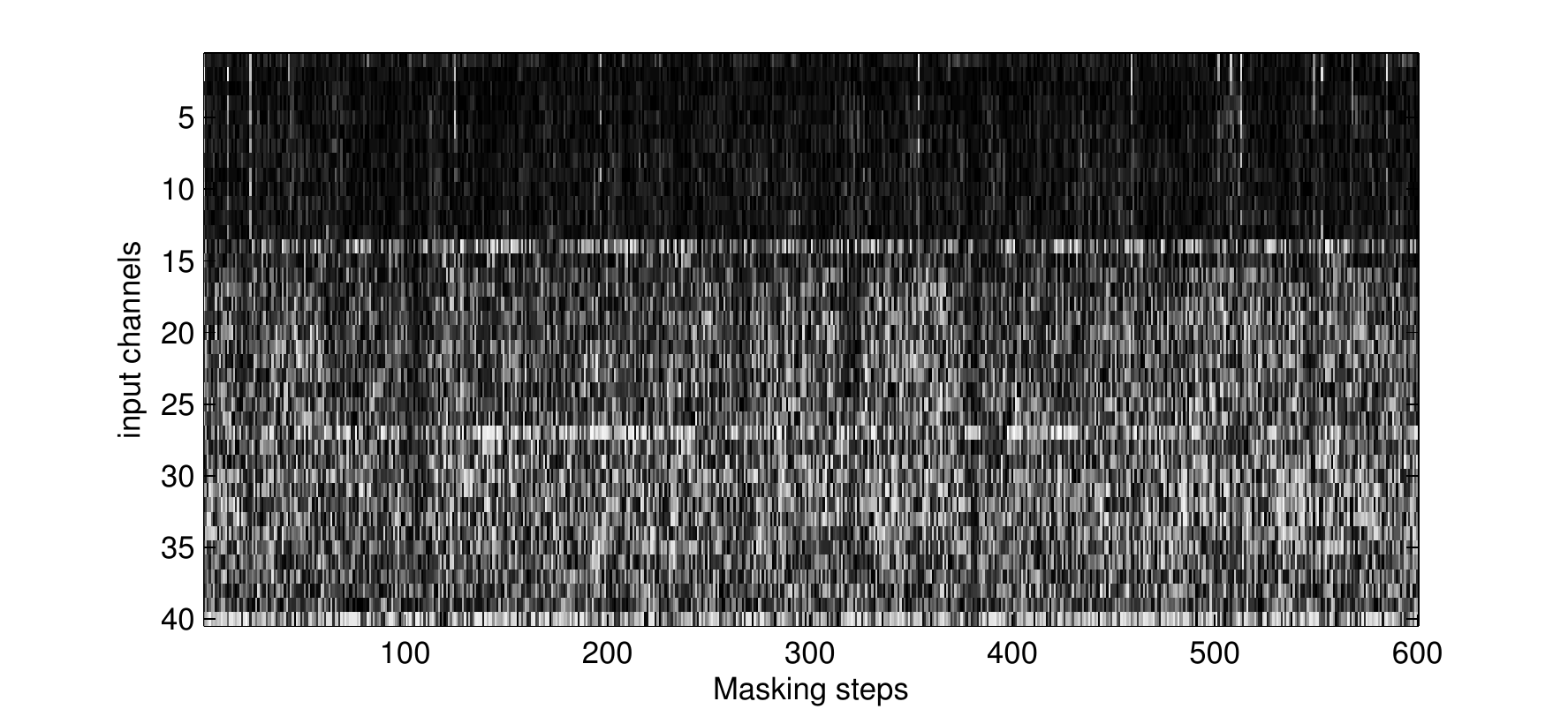}
\caption{Depiction of the input mask trained on the TIMIT task. We have shown the input weights of the 600 masking steps (horizontal axis) for each channel (vertical axis). For the sake of visualization we have here depicted the natural logarithm of the absolute value of the mask plus 0.1. This enhances the difference in scaling for the different channels.}\label{fig:TIMIT_features}
\end{figure}

We tested the use of BPTT for training the input masks both in simulation and in experiment on two benchmark tasks. First, we considered the often-used MNIST written digit recognition dataset, where we use the dynamics of the system indirectly. Next, we applied it on the TIMIT phoneme dataset. For the MNIST experiment we used $N_m = 400$ masking steps. For TIMIT we used $N_m = 600$. 
\subsection{MNIST}
To classify static images using a dynamical system, we follow an approach similar to the one introduced in \cite{Rolfe2013}. Essentially, we repeat the same input segment several times until the state vector $a_i(t)$ of the DCMZ no longer changes. Next we choose the final instance of $a_i(t)$ to classify the image. In practice we used 10 iterations for each image in the MNIST dataset (i.e., each input digit is repeated for 10 masking periods). This sufficed for $a_i(t)$ to no longer depend on its initial conditions, and in practice this meant that we were able to present all digits to the network right after each other.\\
Input masks were trained using  $10^6$ training iterations, where for each iteration the gradient was determined on 500 randomly sampled digits. 
For training we used Nesterov momentum (\cite{Sutskever2013}), with momentum coefficient 0.9, and a learning rate of 0.01 which linearly decayed to zero over the duration of the training. As regularization we only performed 1-pixel shifts for the digits. Note that these 1-pixel shifts were used for training the input masks, but we did not include them when retraining the output weights, as we only presented the DCMZ with the original 60,000 training examples.\\
After training the input weights, we gathered both physical and simulated data for the 4 experiments as described below, and retrained the output weights to obtain a final score. Output weights are trained using the cross-entropy loss function over $10^6$ training iterations, where for each iteration the gradient was determined on 1000 randomly sampled digits. We again used Nesterov momentum, with momentum coefficient 0.9. The learning rate was chosen at 0.002 and linearly decayed to zero. Meta-parameter optimization was performed using 10,000 randomly selected examples from the training set.\\
We performed 4 tests on MNIST. First of all we directly compared performances between the simulated and experimental data. When we visualized the features that the trained input masks generated, we noticed that they seemed ordered (see Figure \ref{fig:MNIST_features}). Indeed, for each masking step, a single set of weights $\mathbf{m}_k$, which can be seen as a receptive field, is applied to the input image, and the resulting signals from the receptive fields are injected into the physical setup one after each other. Apparently, the trained input masks have similar features grouped together in time. To confirm that this ordering in time is a purposeful property, we shuffled the features $\mathbf{m}_k$ over a single masking period to obtain a new input mask without a natural ordering in the features. Next we tested (in simulation) how much the performance degraded when using these masks. Finally, we also tested classification employing masks with completely random elements, where only the scaling of the weights was optimized (which is the RC approach).\\
Results are presented in the middle column of Table \ref{table:Results}. The difference between experimental and simulation results is very small. The time shuffled features do indeed cause a notable increase in the classification error rate, indicating that the optimized input masks actively make use of the internal dynamics of the system, and not just offer a generically good feature set.\\
For the sake of comparison we have added the current state-of-the-art result on MNIST. For a comprehensive overview of results on MNIST please consult \url{http://yann.lecun.com/exdb/mnist/}. Our result are comparable to the best results obtained using neural networks with a  single hidden layer (denoted as a 2-layer NN on the previously mentioned website).

\begin{table}[t]
\begin{center}
\begin{tabular}{rcc}
\hline
 & MNIST test error & TIMIT frame error rate\\
\hline
Experimental data   & 1.16\% &  33.2\%\\
Simulated data &  1.08\%	  & 31.7\% \\
Simulated data: time-shuffled  &1.41\% &   32.8\% \\
Simulated data: random & 6.72\% &  40.5\%  \\
\hline
Best in literature & 0.23\% & 25.0\% \\
& \cite{Ciresan2012}&\cite{Chen2009}\\
\hline
\end{tabular}
\caption{Benchmark performances for different experimental setups.}\label{table:Results}
\end{center}
\end{table}

\subsection{TIMIT}\label{section:TIMIT}
We applied frame-wise phoneme recognition to the TIMIT dataset (\cite{Garofolo1993}). The data was pre-processed to 39-dimensional feature vectors using Mel Frequency Cepstral Coefficients (MFCCs). The data consists of the log energy, and the first 12 MFCC coefficients, enhanced with their first and second derivative (the so-called delta and delta-delta features). The phonemes were clustered into a set of 39 classes, as is the common approach. Note that we did not include the full processing pipeline to include segmentation of the labels and arrive at a phoneme error rate. Here, we wish to illustrate the potential of our approach and demonstrate how realizations of physical computers can be extended to further concepts, rather than to claim state-of-the-art performance. Given that, in addition, the input masks are trained to perform frame-wise phoneme classification, including the whole processing pipeline would not be informative.\\
Input masks are trained using  $50,000$ training iterations, where for each iteration the gradient was determined on 200 randomly sampled sequences of a length of 50 frames. For training we again used Nesterov momentum, with momentum coefficient 0.9, and a learning rate of 0.2 which linearly decayed to zero over the duration of the training. As we were in a regime far from overfitting, we simply chose the training error for meta-parameter optimization. We have depicted the optimized input mask in Figure \ref{fig:TIMIT_features}. Note that the training process strongly rescaled the masking weights for different input channels, putting more emphasis on the delta and delta-delta features (respectively channels 14 to 26 and 27 to 39 ). \\
We repeated the four scenarios previously discussed: using optimized masks in simulation and experiment, using time-shuffled masks, and using random masks. The resulting frame error rates are presented in the right column of Table \ref{table:Results}. The simulated and experimental data differ by 1.5\%, a relatively small difference, indicating that input masks optimized in simulation are useful in practice, even in the presence of unavoidable discrepancies between the used model and the DCMZ. Results for random masks are significantly worse than those with optimized input masks.
\\
Comparison to literature is not straightforward as most publications do not mention frame error rate, but rather the error rate after segmentation. We included the lowest frame error rate mentioned in literature to our knowledge, though it should be stated that other works may have even lower values, even when they are not explicitly mentioned. For an overview of other results on frame error rate please check \cite{Keshet2011}.
\\
The decrease in performance when using time-shuffled masks is quite modest, suggesting that in this case, most of the improvement over random masks is due directly from the features themselves, and the precise details of the dynamics of the system are less crucial than was the case in the MNIST task \footnote{Note that, when the features are shuffled in time over a single masking period, this indirectly also affects the way information is passed on between different masking periods as the communication between specific nodes between masking periods is a combined effect of the low-pass filter and the self connection.}. Although further testing is needed, we suggest two possible reasons for this. First of all, the TIMIT dataset we used contained the first and second derivatives of the first thirteen channels, which already provides information on the preceding and future values and acts as an effective time window. Indeed as can be seen from Figure \ref{fig:TIMIT_features}, the input features amplify these derivatives. Therefore, a lot of temporal context is already embedded in a single input frame, reducing the need for recurrent connections. Secondly, the lack of control over the way information is mixed over time may still pose an important obstacle to effectively use the recurrence in the system.  Currently, input features are trained to perform two tasks at once: provide a good representation of the current input, and at the same time design the features in such a way that they can make use of the (fixed) dynamics present within the system. It may prove the case that the input masks do not have enough modeling power to fulfill both tasks at once, or that the way temporal mixing occurs in the network cannot be effectively put to use for this particular task.

\section{Discussion and Future Work}
In this paper we presented an experimental survey of the use of backpropagation through time on a physical delay-coupled electro-optical dynamical system, in order to use it as a machine learning model. We have shown that such a physical setup can be approached as a special case of recurrent neural network, and consequently can be trained with gradient descent using backpropagation. Specifically, we have shown that both the input and output encodings (input and output \emph{masks}) for such a system can be fully optimized in this way, and that the encodings can be successfully applied to the real physical setup. 
\\
Previous research in the usage of electro-optical dynamical systems for signal processing used random input encodings, which are quite inefficient in scenarios where the input dimensionality is high. We focused on two tasks with a relatively high input dimensionality: the MNIST written digit recognition dataset and the TIMIT phoneme recognition dataset. We showed that in both cases, optimizing the input encoding provides a significant performance boost over random masks. We also showed that the input encoding for the MNIST dataset seems to directly utilize the inherent dynamics of the system, and hence does more than simply provide a useful feature set.
\\
Note that the comparison with Reservoir Computing is based on the constraints by a given physical setup and a given set of resources. We note that the Reservoir Computing setup could give good results on the proposed tasks too, if we were greatly scaling up its effective dimensionality. This has been evidenced in, for example, \cite{Triefenbach2010}, where good results on the TIMIT dataset were achieved by using Echo State Networks (a particular kind of Reservoir Computing) of up to 20,000 nodes. In our setup this would be achieved by increasing the number of masking steps $N_m$ within one masking period. In reality, however, we will face two practical problems. First of all, there are bandwidth limitations in signal generation and measurement. Parts of the signal that fluctuate rapidly would be lost when reducing the duration of a single masking step. 
If one would scale up by keeping the length of the mask steps fixed but use a longer physical delay, for instance a fiber of tens or hundreds of kilometers, the potential gain in performance comes at the cost of one of the systemÕs important advantages: its speed. Also it is hard to foresee how other optical effects in such long fibers such as dispersion and attenuation, would affect performance. This would be an interesting research topic for future investigations.
\\
At the current stage we did not quantify how much the results in this paper hinge on the ability to model the system mathematically. This particular system can be modeled rather precisely, but it is unclear how fast the usefulness of the presented approach would degrade when the model becomes less acute.  
\\
Several directions for future improvements are apparent. The most obvious one is that we could greatly simplify the training process by putting the DCMZ measurements directly in the training loop: instead of optimizing input masks in simulations, we could just as well directly use real, measured data. The Jacobians required for the backpropagation phase can be computed from the measured data. A training iteration would then consist of the following steps: sample data, produce the corresponding input to the DCMZ with the current input mask, measure the output, perform backpropagation in simulation, and update the parameters. The benefit would be that we directly end up with functional input and output masks, without the need for retraining. On top of that, data collection would be much faster. The only additional requirement for this setup would be the need for a single computer controlling both signal generation and signal measurement.
\\
The next direction for improvement would be to rethink the design of the system from a machine learning perspective. The current physical setup on which we applied backpropagation finds its origins in reservoir computing research. As we argue in Section \ref{section:Physical_System}, the system can be considered as a special case of recurrent network with a fixed, specific connection matrix between the hidden states at different time steps. In the reservoir computing paradigm, one always uses fixed dynamical systems that remain largely unoptimized, such that in the past this fact was not particularly restrictive. However, given the possibility of fully optimizing the system that was demonstrated in this paper, the question on how to redesign this system such that we can assert more control over the recurrent connection matrix, and hence the dynamics of the system itself, becomes far more relevant. Currently we have a fixed dynamical system of which we optimize the input signal to accommodate a certain signal processing task. As explained at the end of Section \ref{section:TIMIT}, it appears that backpropagation can currently only leverage the recurrence of the system to a limited degree, when using a single delay loop. Therefore it would be more desirable to optimize both the input signal and the internal dynamics of the system to accommodate a certain task. Alternatively, the configuration can be easily extended to multiple delay loops, allowing for a richer recurrent connectivity. 
\\
The most significant result of this paper is that we have shown experimentally that the backpropagation algorithm, a highly abstract machine learning algorithm, can be used as a tool in designing analog hardware to perform signal processing. This means that we may be able to vastly broaden the scope of research into physical and analog realizations of neural architectures. In the end this may result in systems that combine the best of both worlds: powerful processing capabilities at a tremendous speed and with a very low power consumption.
\section*{Acknowledgements}
P.B., M.H. and J.D. acknowledge support by the interuniversity attraction pole (IAP) Photonics@be of the Belgian Science Policy Office, the ERC NaResCo Starting grant and the European Union Seventh Framework Programme under grant agreement no. 604102 (Human Brain Project). M.C.S. and I.F. acknowledge support by MINECO (Spain), Comunitat Aut\`onoma de les Illes Balears, FEDER, and the European Commission under Projects TEC2012-36335 (TRIPHOP), and Grups Competitius. M.H. and I.F. acknowledge support from the Universitat de les Illes Balears for an Invited Young Researcher Grant. In addition, we acknowledge Prof. L. Larger for developing the optoelectronic delay setup.

\end{document}